\documentclass[final, times, a4paper]{amsart}

\usepackage[left=1.75cm, right=1.75cm, top=2cm, bottom=2cm]{geometry}
\usepackage{natbib}
\usepackage{bbm}
\usepackage{algorithm}
\usepackage{multirow}
\usepackage{paralist}
\usepackage{multicol}
\usepackage{hyperref}
\usepackage{pgfplots}
\usepackage{amsmath}
\usepackage{amsfonts}
\DeclareUnicodeCharacter{2212}{−}
\usepgfplotslibrary{groupplots}
\pgfplotsset{compat=newest}

\DeclareMathOperator*{\argmin}{arg\,min}

\title[Trinary decision trees]{Trinary Decision Trees for handling missing data}
\author[Henning Zakrisson]{Henning Zakrisson \\ \smallskip {\tiny
        Department of Mathematics, Stockholm University}}
\begin{document}

\begin{abstract}
        This paper introduces the Trinary decision tree, an algorithm designed to improve the handling of missing data in decision tree regressors and classifiers.
        Unlike other approaches, the Trinary decision tree does not assume that missing values contain any information about the response.
        Both theoretical calculations on estimator bias and numerical illustrations using real data sets are presented to compare its performance with established algorithms in different missing data scenarios (Missing Completely at Random (MCAR), and Informative Missingness (IM)).
        Notably, the Trinary tree outperforms its peers in MCAR settings, especially when data is only missing out-of-sample, while lacking behind in IM settings.
        A hybrid model, the TrinaryMIA tree, which combines the Trinary tree and the Missing In Attributes (MIA) approach, shows robust performance in all types of missingness.
        Despite the potential drawback of slower training speed, the Trinary tree offers a promising and more accurate method of handling missing data in decision tree algorithms.

\end{abstract}

\maketitle

\begin{multicols}{2}
\raggedcolumns
\section{Introduction} \label{sec:intro}
        Missing values are prevalent in real data.
        As noted by e.g.\ ~\cite{nijman2022missing}, this is often not handled or mentioned in machine learning applications in a satisfactory way.
        Classification and Regression Trees (CART), as defined by ~\cite{breiman1984classification} provide numerous ways to handle missing values in covariates.
        Since CARTs are the foundation of many increasingly popular machine learning algorithms such as Gradient Boosting Machines (GBMs)~\citep{friedman2001greedy}, Random Forests~\citep{ho1995random}, and XGBoost~\citep{chen2015xgboost}, they are still relevant today.
        But the proposed methods of handling missing data come with drawbacks.

        The simplest way to handle missing values when training a tree is to simply ignore them by discarding data points with any missing feature.
        This of course means losing potentially useful information, and is not an option when predicting using data with missing values.
        The perhaps second-simplest method is using the \textit{majority rule}, where data points with missing values are assigned to the category in a tree split with the largest amount of data in training.
        Another method is presented by ~\cite{twala2008good}, the Missing In Attributes (MIA) algorithm.
        MIA assigns data points with missing covariates to the category that minimizes the loss for the training data.
        This is similar, but not identical, to assigning missing values an own category in a categorical feature.
        ~\cite{quinlan1993c4} introduces the C4.5 algorithm for decision trees, and with that a weighted probabilistic strategy for missing value-handling, henceforth referred to as Fractional Case - FC.
        In FC, a data point with a missing value in a split is assigned a weight of membership in both categories of a binary splot, with the weight depending on the distribution of the observable data in the node.
        For out-of-sample data, the weights for all terminal nodes are calculated and the prediction is given as a weighted average.
        ~\cite{breiman1984classification} proposes using so-called \textit{surrogate splits} in order to find other covariates on which the data points which lack the relevant observation can be split to form similar splits.
        This requires that there are no missing values in the surrogate covariate - or that a secondary surrogate variable is found in its place.

        Cons of these methods include losing potentially useful information (discarding data, FC), assuming there is always information in missingness (MIA), requiring missing values in the training data to be able to handle missing values in out-of-sample prediction (MIA, surrogate splits) or losing interpretability (FC).

        The trinary decision tree for missing value handling (henceforth Trinary tree) introduced in this paper has four important attributes:
        \begin{compactitem}
        \item It does not assume that missing data points contain any information about the response
        \item It can handle missing values in predictions even if it was trained on a data set with no missing data
        \item It maintains the interpretability of a standard decision tree
        \item It produces locally unbiased estimators of tree node values - which the other algorithms do not necessarily do
        \end{compactitem}

        The first three are apparent from the algorithm, which is presented in Section ~\ref{sec:algorithm}, whereas the fourth one is proven in Section ~\ref{sec:bias}.
        In Section ~\ref{sec:numerical}, the algorithm is tested against its peers with real data sets.

\section{The Trinary tree} \label{sec:algorithm}
        Consider a loss function $\mathcal{L}((y_i)_{i\in \mathcal{I}}, \delta)$, where $\mathcal{I}$ is an index set and $\delta$ is a parameter.
        For regression problems, $\delta$ is a real number, and for classification problems, $\delta$ is a probability vector.
        In this paper, two loss functions will be considered: the sum of squared errors (SSE) for regression and the point-wise cross-entropy for classification.
        These are defined as
        \begin{equation}
        \mathcal{L}^{\textrm{SSE}}((y_i)_{i\in \mathcal{I}}, \delta) = \sum_{i\in \mathcal{I}} (y_i - \delta)^2,
        \end{equation}
        and
        \begin{equation}
        \mathcal{L}^{\textrm{XE}}((y_i)_{i\in \mathcal{I}}, \delta) = \sum_{i\in \mathcal{I}} \log(\delta_{y_i})
        \label{eq:loss}
        \end{equation}
        respectively.
        Let $\delta_{\mathcal{I}}$ denote the minimizer of $\mathcal{L}((y_i)_{i\in \mathcal{I}}, \delta)$, i.e.\
        \begin{equation}
        \delta_{\mathcal{I}} = \argmin_{\delta} \mathcal{L}((y_i)_{i\in \mathcal{I}}, \delta)
        \end{equation}
        For both SSE and point-wise cross-entropy, $\delta_{\mathcal{I}}$ has a closed form solution, namely
        \begin{equation}
        \delta_{\mathcal{I}}^{\textrm{SSE}} = \frac{1}{|\mathcal{I}|} \sum_{i\in \mathcal{I}} y_i
        \end{equation}
        and
        \begin{equation}
        \left(\delta_{\mathcal{I}}^{\textrm{XE}}\right)_k = \frac{1}{|\mathcal{I}|} \sum_{i\in \mathcal{I}} \mathbbm{1}_{\{y_i\}=k}, \quad k=1,\dots,K
        \label{eq:delta_estimates}
        \end{equation}
        where $K$ is the number of classes in the classification problem.
        A binary decision tree is generally fitted greedily by minimizing the loss function at each so called \textit{node} separately, starting with the \textit{root node} containing all data points.
        For data set $(y_i,x_i)_{i=1}^n$, where $y_i \in \mathcal{Y}$ is the response and $x_{ij} \in \mathcal{X}_j$, $j=1,\dots,p$, is a covariate, this means finding a combination of a covariate $j$ and covariate subspaces $\mathcal{X}_j^l,\mathcal{X}_j^r \subset \mathcal{X}_j$ that minimize
        \begin{equation}
        \mathcal{L}(
        (y_i)_{
                i \in \mathcal{I}_{jl}
        },
        \delta_
                {
                \mathcal{I}_{jl}
                }
        ) + \mathcal{L}((y_i)_{
                i \in
                \mathcal{I}_{jr}
        }, \delta_{
                \mathcal{I}_{jr}
                }
        ) ,
        \label{eq:split}
        \end{equation}
        where
        \begin{equation}
        \mathcal{I}_{jl} = \{i \in \mathcal{I} : x_{ij} \in \mathcal{X}_j^l\},
        \qquad
        \mathcal{I}_{jr} = \{i \in \mathcal{I} : x_{ij} \in \mathcal{X}_j^r\},
        \label{eq:index_sets}
        \end{equation}
        such that $\mathcal{X}_j^l \cup \mathcal{X}_j^r = \mathcal{X}_j$.
        In the case where $\mathcal{X} \in \mathbb{R}$, $\mathcal{X}_j^l$ and $\mathcal{X}_j^r$ are constrained to be continuous intervals.
        After finding the optimal split, the procedure is repeated for the so called \textit{daughter nodes}, i.e.\ the nodes that contain the data points from the two sides of the split.
        The procedure is generally continued until reaching some stopping criterion, such as a maximum tree depth.
        The nodes that are not split are called \textit{terminal nodes}.

        Data points where the chosen splitting covariate is missing can be handled in a number of ways.
        The \textit{MIA} strategy assigns them to the daughter node that provides the largest reduction in the loss function (see Algorithm~\ref{alg:mia} in Appendix~\ref{sec: algorithms}).
        The \textit{majority} strategy assigns them to the daughter node with the largest amount of data (see Algorithm~\ref{alg:majority} in Appendix~\ref{sec: algorithms}).
        The \textit{FC} strategy assigns them to both daughter nodes with weights depending on the distribution of the observable data in the node (see Algorithm~\ref{alg:fc} in Appendix~\ref{sec: algorithms}).

        In contrast, the \textit{Trinary} strategy assigns them to a third daughter node, which changes the function to minimize in~\eqref{eq:split} to
        \begin{equation}
        \mathcal{L}(
        (y_i)_{
        i \in \mathcal{I}_{jl}
        },
        \delta_
                {
        \mathcal{I}_{jl}
                }
        ) + \mathcal{L}((y_i)_{
        i \in \mathcal{I}_{jr}
        }, \delta_{
        \mathcal{I}_{jr}
                }
        )
        + \mathcal{L}((y_i)_{
                i \in \mathcal{I}_{jm}
        }, \delta_{
                \mathcal{I}
                }
        ),
        \label{eq:trinary_split}
        \end{equation}
        where $\mathcal{I}_{jm} = \{i \in \mathcal{I} : x_{ij} \textrm{ missing}\}$.
        Note that $\delta_{\mathcal{I}}$ in the third term is the minimizer of the loss function over the entire data set of the node.
        This means that the third term evaluates the loss of the points that are not assigned to the left or right nodes as if it retained the parameter estimate of the mother node.
        After finding a split, the procedure is repeated for all three daughter nodes.
        For the third node, the entire data set is used for further splitting, but omitting the splitting covariate.
        Thus, the third node will grow further by first splitting on the second-best covariate, then continue to grow.
        The third node is considered to be at the same depth level as the mother node, since the data set has not been split.

        The point of the third node is to avoid making assumptions about the missing data.
        By not assigning the missing data to either of the standard daughter nodes, the Trinary tree does not contaminate the $\delta$ estimates for the standard nodes with data points that do not belong there.
        Instead, the Trinary tree uses the entire data set to estimate $\delta$ for the third node, as a way to regularize predictions when in uncertainty about important covariates.

\begin{figure*}[ht]
        \centering
        \usetikzlibrary{positioning}

\begin{tikzpicture}[
  every node/.style={
    circle,
    draw,
    fill=blue!20,
    align=center,
  },
  edge from parent/.style={draw, ->},
]
\node (root) {$\delta_0$}
  child {
    node (left_child)  {$\delta_1$}
    edge from parent node[
    draw = none,
    fill = none,
    left,
    ] {$x_{i1} \in \mathcal{X}_1^l$}
  }
  child{
    node (right_child)  {$\delta_2$}
  edge from parent node[
      draw = none,
      fill = none,
      right,
      ] {$x_{i1} \in \mathcal{X}_1^r$}
  };

\node[right=4cm of root] (missing_child) {$\delta_0$}
  child {
    node (missing_left_child)  {$\delta_3$}
    edge from parent node[
      draw = none,
      fill = none,
      left,
      ] {$x_{i2} \in \mathcal{X}_2^l$}
  }
  child{
    node (missing_right_child)  {$\delta_4$}
    edge from parent node[
      draw = none,
      fill = none,
      right,
      ] {$x_{i2} \in \mathcal{X}_2^r$}
  };

\draw[dotted, ->] (root) -- (missing_child) coordinate[midway] (missing_edge);
\node[
    above=-0.75cm of missing_edge,
    draw = none,
    fill = none,
] (label) {$x_{i1} \textrm{ missing}$};

\node[right=3cm of missing_child] (missing_child_2) {$\delta_0$};
\draw[dotted, ->] (missing_child) -- (missing_child_2) coordinate[midway] (missing_edge_2);
\node[
    above=-0.75cm of missing_edge_2,
    draw = none,
    fill = none,
] (label) {$x_{i2} \textrm{ missing}$};

\end{tikzpicture}
        \caption{Visualization of a Trinary tree with depth $1$ for a covariate with $p=2$ dimensions.
        Note that since the third node is considered to be at the same depth level as the root node, an additional split is made.
        Since the best performing split covariate $j=1$ is no longer available, the second-best split covariate $j=2$ is used for the second split.
        Since that covariate could also be missing, the third node has its own third daughter node.
        Since there are no further covariates to split on, that node is a terminal node.
        }
        \label{fig:trinary_tree}
        \end{figure*}
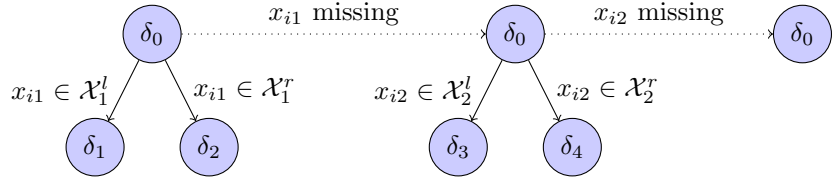

        The Trinary tree training algorithm is summarized in Algorithm ~\ref{alg:trinary}.
        A visualization of a Trinary tree with depth $1$ is shown in Figure ~\ref{fig:trinary_tree}.

        \begin{algorithm}[H] 
                \caption{Trinary tree training algorithm}
                \label{alg:trinary}
                \begin{flushleft}
                        \textbf{Let}
                        \begin{compactitem}
                        \item $(y_i,x_i)_{i\in \mathcal{I}}$, where $y_i \in \mathcal{Y}$, $x_{ij} \in \mathcal{X}_j$, $j = 1,\ldots,p$, be the training data
                        \item $\mathcal{L}((y_i)_{i \in \mathcal{I}},\delta)$ be the loss given parameter $\delta$
                        \item $\delta_{\mathcal{I}}$ be the minimizing parameter of $\mathcal{L}((y_i)_{i \in \mathcal{I}},\delta)$
                        \item $\mathcal{I}_{jl} = \{i \in \mathcal{I}: x_{ij} \in \mathcal{X}_j^{l}\}$, $\mathcal{I}_{jr} = \{i \in \mathcal{I}: x_{ij} \in \mathcal{X}_j^{r}\}$
                        \item $\mathcal{I}_{jm} = \{i: x_{ij} \textrm{ missing}\}$
                        \item $d_{\textrm{max}}$ be the maximum depth
                        \item $n$ be the minimum number of samples per node
                        \end{compactitem}
                        \textbf{Define training function} $\mathcal{T}\left((y_i,x_i)_{i \in \mathcal{I}},d\right)$ $\rightarrow$ $h(x)$:
                        \begin{compactitem}
                        \item[] \textbf{If} $d=d_{\textrm{max}}$ or $|\mathcal{I}| = n$:
                        \begin{compactitem}
                                \item[] \textbf{Output}
                                \begin{align*}
                                h(x) = \delta_\mathcal{I}
                                \end{align*}
                        \end{compactitem}
                        \item[] \textbf{Else}:
                        \begin{compactitem}
                                \item[] \textbf{Fit}
                        \begin{compactitem}
                        \item[] Find $j$, $\mathcal{X}_j^{l}$, and $\mathcal{X}_j^{r}$ that minimize
                                \[
                                \hspace{1cm}
                                \mathcal{L}\left(
                                (y_i)_{i \in \mathcal{I}_{jl}},\delta_{\mathcal{I}_{jl}}
                                \right)
                                \! + \!
                                \mathcal{L}\left(
                                (y_i)_{i \in \mathcal{I}_{jr}},\delta_{\mathcal{I}_{jr}}
                                \right)
                                \! + \!
                                \mathcal{L}\left(
                                (y_i)_{i \in \mathcal{I}_{jm}},\delta_{\mathcal{I}}
                                \right)
                                \]
                                \item[] such that $\left| \mathcal{I}_l  \right| \geq n$, $\left| \mathcal{I}_r  \right| \geq n$, $\mathcal{X}_j^l \cup \mathcal{X}_j^r = \mathcal{X}_j$
                        \end{compactitem}
                        \item[] \textbf{Grow}
                                \begin{align*}
                                        h_l(x) & = \mathcal{T}\left(\left(y_i,x_i\right)_{\mathcal{I}_l},d+1 \right) \\
                                        h_r(x) & = \mathcal{T}\left(\left(y_i,x_i\right)_{\mathcal{I}_r},d+1 \right) \\
                                        h_m(x) & = \mathcal{T}\left(\left(y_i,x_i\right)_{\mathcal{I}},d \right)
                                \end{align*}
                        \item[] \textbf{Output}
                                \begin{align*}
                                h(x) = \begin{cases}
                                h_l(x), & x_{\cdot j} \in \mathcal{X}_{j}^{l} \\
                                h_r(x), & x_{\cdot j} \in \mathcal{X}_{j}^{r} \\
                                h_m(x), & x_{\cdot j} \textrm{ missing}
                                \end{cases}
                                \end{align*}
                        \end{compactitem}
                        \end{compactitem}
                        \end{flushleft}
        \end{algorithm}

\section{Tree-fitting estimate bias} \label{sec:bias}
        In order to illustrate how the Trinary tree might be preferable to the other methods, let us examine a simple example where the non-trinary methods estimators are locally biased.
        Consider the data set $\mathcal{D} = (X_i,Y_i)_{i=1}^n$, where $X_i \in \mathcal{X}$ is the covariate and $Y_i \in \mathbb{R}$ is the response.
        Let the expected value of $Y$ follow a tree structure, i.e.\  let $\mathbb{E}[Y|X=x] = h(x)$ where
        \[
        h(x) = \begin{cases}
        a, & x \in \mathcal{X}^l \\
        b, & x \in \mathcal{X}^r
        \end{cases}
        \]
        where $\mathcal{X}^l \cup \mathcal{X}^r = \mathcal{X}$, $a<b$ and $\mathbb{P}(X \in \mathcal{X}^r) = p$, $0 < p < 1$.
        Then, consider a censored dataset $\widetilde{\mathcal{D}} = (\tilde{X}_i,Y_i)_{i=1}^n$ where
        \[
        \tilde{X_i} = \begin{cases}
        \texttt{nan}, & \textrm{with probability } q \\
        X_i, & \textrm{with probability } 1-q \\
        \end{cases}
        \]
        where $\texttt{nan}$ corresponds to a missing value, $q>0$.
        Now, consider fitting trees to this for an SSE loss function, using the Majority, MIA and FC algorithms respectively, and examine their estimates of $a$.
        For brevity, let $\mathcal{I}_l$ and $\mathcal{I}_r$ be defined as in ~\eqref{eq:index_sets} and introduce index sets
        \begin{equation*}
        \mathcal{I}^{o} = \left\{i: \tilde{X}_i = X_i\right\},
        \qquad
        \mathcal{I}^{m} = \left\{i: \tilde{X}_i = \texttt{nan}\right\},
        \end{equation*}
        as well as the intersections
        \begin{align*}
        \mathcal{I}^o_l & = \mathcal{I}^o \cap \mathcal{I}_l, &  \mathcal{I}^o_r & = \mathcal{I}^o \cap \mathcal{I}_r, \\
        \mathcal{I}^m_l & = \mathcal{I}^o \cap \mathcal{I}_l, &  \mathcal{I}^m_r & = \mathcal{I}^m \cap \mathcal{I}_r.
        \end{align*}

        For the majority rule algorithm, the estimate of $a$ is
        \[
        \hat{a}_{\textrm{Maj}} = \begin{cases}
        \frac{1}{\left|\mathcal{I}_l^o \right|} \sum \limits_{i \in \mathcal{I}_l^o } Y_i, & \textrm{if } \left| \mathcal{I}_l^o \right| > \left| \mathcal{I}_r^o \right|, \\
        \frac{1}{\left|\mathcal{I}_l^o \right| + \left|\mathcal{I}^m \right|} \sum \limits_{i \in \mathcal{I}_l^o \cup \mathcal{I}^m } Y_i, & \textrm{else.}\\
        \end{cases}
        \]
        First note that
        \[
        \mathbb{E}[\hat{a}_{\textrm{Maj}} | \left| \mathcal{I}_l^o \right| \leq \left| \mathcal{I}_r^o \right|] = a.
        \]
        Also,
        \begin{align*}
        & \mathbb{E}\left[ \hat{a}_{\textrm{Maj}}
        | \left| \mathcal{I}_l^o \right| > \left| \mathcal{I}_r^o \right|
        \right] \\ &
        = \mathbb{E}\left[
        \frac{1}{\left|\mathcal{I}_l^o \right| + \left|\mathcal{I}^m_l \right| + \left|\mathcal{I}^m_r \right|}
        \left(
        \sum \limits_{i \in \mathcal{I}_l^o} Y_i
        + \sum \limits_{i \in \mathcal{I}_l^m} Y_i
        + \sum \limits_{i \in \mathcal{I}_r^m} Y_i
        \right)
        \right] \\ &
        =
        \mathbb{E} \left[
        \frac{1}{\left|\mathcal{I}_l^o \right| + \left|\mathcal{I}^m_l \right| + \left|\mathcal{I}^m_r \right|}
        \left(
        a\left|\mathcal{I}_l^o \right| + a\left|\mathcal{I}^m_l \right| + b\left|\mathcal{I}^m_r \right|
        \right)
        \right]
        \\ &
        \geq
        \frac{1}{\mathbb{E} \left[\left|\mathcal{I}_l^o \right| + \left|\mathcal{I}^m_l \right| + \left|\mathcal{I}^m_r \right|\right]}
        \mathbb{E}\left[
        a\left|\mathcal{I}_l^o \right| + a\left|\mathcal{I}^m_l \right| + b\left|\mathcal{I}^m_r \right|
        \right] \\ &
        = a + \frac{pq}{1-p+pq}\left(b-a\right)
        \end{align*}
        where the concavity of the function
        \begin{align*}
        g(x,y,z) = \frac{x}{x+y+z}
        \end{align*}
        is used for the inequality.
        Then, introduce $\kappa = \mathbb{P}\left(\left| \mathcal{I}_l^o \right| \leq \left| \mathcal{I}_r^o \right|\right)$, and note that
        \begin{align*}
                \mathbb{E}\left[ \hat{a}_{\textrm{Maj}} \right] &
                = \kappa \mathbb{E}[\hat{a}_{\textrm{Maj}} | \left| \mathcal{I}_l^o \right| \leq \left| \mathcal{I}_r^o \right|]
                +
                \left(1-\kappa\right) \mathbb{E}[\hat{a}_{\textrm{Maj}} | \left| \mathcal{I}_l^o \right| > \left| \mathcal{I}_r^o \right|] \\
                & \geq  \kappa a +\left(1-\kappa\right)  \left(a + \frac{pq}{1-p+pq}\left(b-a\right)\right) \\
                & = a +  \frac{\left(1-\kappa\right)pq}{1-p+pq}\left(b-a\right) \\
                & > a
        \end{align*}
        The proof for the MIA strategy is identical with the only change that $\kappa$ then means the probability that the loss is lower if the missing values are assigned to the left node.
        For the Fractional Case strategy the estimate of parameter $a$ has expected value
        \begin{align*}
        \mathbb{E}\left[
        \hat{a}_{\textrm{FC}}
        \right]
        & =
        \mathbb{E}\left[
        \frac{1}{\sum \limits_{i=1}^{n} w_i^l}\sum \limits_{i=1}^{n} w_i^{l} Y_i
        \right] \\
        & \geq \frac{1}{\mathbb{E}\left[\sum \limits_{i=1}^{n} w_i^l\right]}\mathbb{E}\left[\sum \limits_{i=1}^{n} w_i^{l} Y_i
        \right] \\
        & = a + pq\left(b-a\right) \\
        & > a.
        \end{align*}
        where it can also be shown that in order for $\mathbb{E}[\hat{Y}_{\textrm{FC}}] = \mathbb{E}[Y]$ to hold it is required that $\mathbb{E}[\hat{b}_{\textrm{FC}}] < b$.
        Finally, for the Trinary tree, see that
        \begin{align*}
        \mathbb{E}\left[\hat{a}_{\textrm{Tri}} \right]  = \mathbb{E} \left[
        \frac{1}{\left| \mathcal{I}^o_l \right|} \sum \limits_{i \in  \mathcal{I}^o_l} Y_i
        \right]
        = a.
        \end{align*}

\section{Numerical illustration} \label{sec:numerical}
        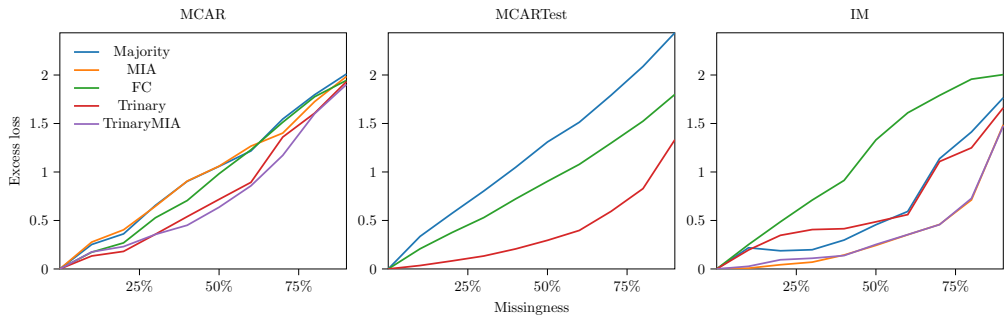
\begin{figure*}[ht]
                \centering
\begin{tikzpicture}[scale=0.55]

\definecolor{crimson2143940}{RGB}{214,39,40}
\definecolor{darkgray176}{RGB}{176,176,176}
\definecolor{darkorange25512714}{RGB}{255,127,14}
\definecolor{forestgreen4416044}{RGB}{44,160,44}
\definecolor{mediumpurple148103189}{RGB}{148,103,189}
\definecolor{steelblue31119180}{RGB}{31,119,180}

\begin{groupplot}[group style={group size=3 by 1}]
\nextgroupplot[
tick align=outside,
tick pos=left,
title={MCAR},
x grid style={darkgray176},
xmin=0, xmax=90,
xtick style={color=black},
xtick={25,50,75},
xticklabels={25\%, 50\%, 75\%},
y grid style={darkgray176},
ylabel={Excess loss},
ymin=0, ymax=2.43517956710787,
ytick style={color=black},
legend style = {at = {(0.03,0.97)}, anchor = north west},
legend style= {draw = none}
]
\addplot [very thick, steelblue31119180]
table {%
0 0
10 0.252452553579306
20 0.361842135555464
30 0.655503864928303
40 0.903922222580627
50 1.05797714578855
60 1.21682740803999
70 1.54613111581882
80 1.79650929722995
90 2.01055269363393
};
\addlegendentry{Majority}
\addplot [very thick, darkorange25512714]
table {%
0 0
10 0.277135156096776
20 0.403482646893317
30 0.646336876808034
40 0.906285457300083
50 1.0602887865803
60 1.26652882526516
70 1.40114314794608
80 1.72552033052569
90 1.98507606962195
};
\addlegendentry{MIA}
\addplot [very thick, forestgreen4416044]
table {%
0 0
10 0.173091071582008
20 0.269005345916281
30 0.52786229841692
40 0.705676060024672
50 0.982170564842401
60 1.2283297761153
70 1.51312446044245
80 1.775939193026
90 1.94461134690981
};
\addlegendentry{FC}
\addplot [very thick, crimson2143940]
table {%
0 0
10 0.134041936342174
20 0.18108013670154
30 0.35791716836869
40 0.539880573102399
50 0.717793232348596
60 0.893069202790649
70 1.3602199832579
80 1.60672961849593
90 1.93186174515328
};
\addlegendentry{Trinary}
\addplot [very thick, mediumpurple148103189]
table {%
0 0
10 0.175634342014173
20 0.23079410741231
30 0.356417786123386
40 0.450913144601945
50 0.636453141212769
60 0.857330381026772
70 1.17134166515981
80 1.60086346542038
90 1.90189571083013
};
\addlegendentry{TrinaryMIA}

\nextgroupplot[
tick align=outside,
tick pos=left,
title={MCARTest},
x grid style={darkgray176},
xlabel={Missingness},
xmin=0, xmax=90,
xtick style={color=black},
xtick={25,50,75},
xticklabels={25\%, 50\%, 75\%},
y grid style={darkgray176},
ymin=0, ymax=2.43517956710787,
ytick style={color=black}
]
\addplot [very thick, steelblue31119180]
table {%
0 0
10 0.335654668324768
20 0.57271990801076
30 0.803337244844399
40 1.04762892079803
50 1.30915816297445
60 1.51252960584904
70 1.79219808552696
80 2.0878432695503
90 2.43517956710787
};
\addplot [very thick, forestgreen4416044]
table {%
0 0
10 0.207664802576199
20 0.375005704506372
30 0.529451673315216
40 0.721973301336609
50 0.903216404205325
60 1.0784401412091
70 1.29835487926756
80 1.52349160829779
90 1.79939334871536
};
\addplot [very thick, crimson2143940]
table {%
0 0
10 0.035220022476302
20 0.0827209543080682
30 0.133296759132036
40 0.206365399346567
50 0.295056201404468
60 0.396973359751501
70 0.594207117761779
80 0.827998752430242
90 1.33166014696032
};

\nextgroupplot[
tick align=outside,
tick pos=left,
title={IM},
x grid style={darkgray176},
xmin=0, xmax=90,
xtick style={color=black},
xtick={25,50,75},
xticklabels={25\%, 50\%, 75\%},
y grid style={darkgray176},
ymin=0, ymax=2.43517956710787,
ytick style={color=black}
]
\addplot [very thick, steelblue31119180]
table {%
0 0
10 0.218472611758376
20 0.187422418346787
30 0.198174117754783
40 0.297932651380951
50 0.456002379699301
60 0.593156376469837
70 1.13762929345747
80 1.41251100153834
90 1.76519144967589
};
\addplot [very thick, darkorange25512714]
table {%
0 0
10 0.00793815716035454
20 0.0433056446127727
30 0.0703781094483131
40 0.143851444422331
50 0.242981045053142
60 0.352752337129343
70 0.458820170037927
80 0.712035666101393
90 1.48704364887424
};
\addplot [very thick, forestgreen4416044]
table {%
0 0
10 0.251404975800326
20 0.485996721634159
30 0.710308520664267
40 0.912993269809949
50 1.3304357341759
60 1.61019501460336
70 1.78981915777359
80 1.95710024808507
90 2.00364534211667
};
\addplot [very thick, crimson2143940]
table {%
0 0
10 0.195685196463572
20 0.346790489726822
30 0.406070207006752
40 0.4148868360563
50 0.485900514411761
60 0.560504151840329
70 1.10803485154561
80 1.248975393771
90 1.66166310189262
};
\addplot [very thick, mediumpurple148103189]
table {%
0 0
10 0.0271385844719515
20 0.0947373656086724
30 0.110232083703675
40 0.137888625063287
50 0.252295813085689
60 0.354182870839338
70 0.457455288287488
80 0.727682388084242
90 1.47564823733976
};
\end{groupplot}

\end{tikzpicture}
                \caption{Average excess loss per missingness ratio for the tree algorithms in different kinds of missingness on all data sets}
                \label{fig:results}
        \end{figure*}
        In order to illustrate the benefits of the Trinary tree, data with increasing missingness is created for the data sets in Table~\ref{tab:data_sets}.
        All the data sets are available online.
        The data sets have been chosen in order to provide a wide array of applications, and varying characteristics of the data.
        The performance of the individual data sets will not be evaluated specifically in this paper, but rather the performance of the algorithms on the data sets as a whole.
        The data sets have been minimally pre-processed, e.g.\ by removing any potential missing values.
        This is done in order to have full control over the missingness.
        The tree depth is first tuned by performing $10$-fold cross validation on the full data set, with a maximum tree depth set to $5$.
        For the classification problems, the folds are stratified so that the relative class frequency is equal in every fold.
        The dataset is then censored, i.e.\  values are replaced with missing values, (with missingness $q\%$ ranging from $0\%$ to $90\%$) in three different ways.
        \begin{itemize}
        \item[\textbf{MCAR}] $q\%$ of the data is removed from all features in the training and test set, completely at random.
        \item[\textbf{MCARTest}] $q\%$ of the data is removed from all features in the test set, completely at random.
        The training set has no missing values.
        \item[\textbf{IM}] $q\%$ of the data is removed from all features in the training and test set.
        For numerical features, the largest values are removed first.
        For categorical ones, they are removed on a category-by-category basis.
        \end{itemize}
        Thereafter, the four different tree algorithms (Majority, MIA, FC, and Trinary) are trained and evaluated on the $10$ folds.
        Additionally, a fifth tree algorithm, denoted TrinaryMIA is evaluated.
        This is an amalgamation of Trinary and MIA that in every node evaluates whether a MIA-style or Trinary Tree style split reduces the training loss the most and then picks the one that does.
        \begin{table}[H] \label{tab:data_sets}
                \centering
                \caption{Data sets}
                \begin{tabular}{c|lrr}
                        type &  Name & size & features   \\ \hline
                \multirow{4}{*}{regression} & AutoMPG & 392 & 8 \\
                        & Black Friday & 550,068 & 6 \\
                        & Cement & 1,030 & 9 \\
                        & Life Expectancy & 138 & 17 \\ \hline
                \multirow{4}{*}{classification}  & Titanic & 712 & 7\\
                        & Lymphography & 142 & 19 \\
                        & Boston Housing & 506 & 14 \\
                        & Wheat seeds & 199 & 8
                \end{tabular}
        \end{table}
        First the total loss is calculated for the entire data set with no missing values.
        Then the total loss is calculated for the data set with increasing missingness, and the excess loss (i.e.\  the loss divided by the loss in the case with no missing values) is calculated.
        The average excess loss over all data sets is presented in Figure ~\ref{fig:results}.
        Since the MIA and majority strategies are identical in cases where there is no missing data in the training data, MIA is omitted from the middle figure.
        The same applies to Trinary and TrinaryMIA trees.

        As can be seen, the TrinaryMIA tree is the best performer in the MCAR case, followed by the Trinary tree.
        For higher levels of missingness all algorithms seem to perform almost equally bad.
        In the MCARTest case, the Trinary tree is the best performer, followed by the FC tree.
        For the IM case, the TrinaryMIA and MIA trees perform very similarly, and are the best performers.
        The performance of MIA in this setting is expected, but it seems like the TrinaryMIA tree is able to find the appropriate splits as well.

\section{Concluding remarks} \label{sec:conclusion}
        It is clear both from the bias calculations and the numerical illustration that the Trinary tree has benefits over its peers in MCAR settings.
        Especially, the performance of the algorithm in the case where data is only missing in the test set is noteworthy.
        It is however important to remember that assuming that missingness contains no information is an assumption in itself - seen by the less impressive performance in the IM test.
        This drawback seems to be easily overcome by the TrinaryMIA tree, which maintains performance in all types of missingness.
        Surprisingly, the TrinaryMIA tree also outperforms the Trinary tree in the MCAR case, however not by a large margin.

        The potential of using the Trinary tree algorithm as a weak learner in more powerful machine learning algorithms, such as a GBM, is an interesting topic, since the reguralization and missing value-handling would then be inherited by the ensemble model.

        A drawback for the Trinary tree, is that for large data sets (especially covariate data sets with many features and categorical features with high cardinality) or deep trees, the training speed can suffer.
        For shallow trees and data sets with a limited number of covariates, the speed is however on par with the other methods.
        It should also be noted that a large part of the tree training can be parallelized, since nodes can be trained independently of each other when using standard greedy splitting.
        Also, TrinaryMIA training is often faster since it, if there is information in missingness in the training data, will grow fewer nodes than a standard Trinary tree.

\section*{Acknowledgements}
        The author would like to thank Mathias Lindholm for his valuable comments and suggestions.
        The trinary tree algorithm is available in the \texttt{python} package \texttt{trinary-tree} on \texttt{PyPI},
        with the source code available at \url{https://github.com/henningzakrisson/trinary-tree}.
        Note that the package is a proof-of-concept implementation and is not optimized for speed.

\bibliographystyle{agsm}
\bibliography{refs}
\columnbreak
\appendix

\section{Tree algorithms} \label{sec: algorithms}

        \begin{algorithm}[H] 
                \caption{Missing In Attributes training algorithm}
                \label{alg:mia}
                \begin{flushleft}
                        \textbf{Let}
                        \begin{compactitem}
                        \item $(y_i,x_i)_{i\in \mathcal{I}}$, where $y_i \in \mathcal{Y}$, $x_{ij} \in \mathcal{X}_j$, $j = 1,\ldots,p$, be the training data
                        \item $\mathcal{L}((y_i)_{i \in \mathcal{I}},\delta)$ be the loss given parameter $\delta$
                        \item $\delta_{\mathcal{I}}$ be the minimizing parameter of $\mathcal{L}((y_i)_{i \in \mathcal{I}},\delta)$
                        \item $\mathcal{I}_{jl} = \{i \in \mathcal{I}: x_{ij} \in \mathcal{X}_j^{l}\}$, $\mathcal{I}_{jr} = \{i \in \mathcal{I}: x_{ij} \in \mathcal{X}_j^{r}\}$
                        \item $d_{\textrm{max}}$ be the maximum depth
                        \item $n$ be the minimum number of samples per node
                        \end{compactitem}
                        \textbf{Define training function} $\mathcal{T}\left((y_i,x_i)_{i \in \mathcal{I}},d\right)$ $\rightarrow$ $h(x)$:
                        \begin{compactitem}
                        \item[] \textbf{If} $d=d_{\textrm{max}}$ or $|\mathcal{I}| = n$:
                        \begin{compactitem}
                                \item[] \textbf{Output}
                                \begin{align*}
                                h(x) = \delta_\mathcal{I}
                                \end{align*}
                        \end{compactitem}
                        \item[] \textbf{Else}:
                        \begin{compactitem}
                                \item[] \textbf{Fit}
                        \begin{compactitem}
                        \item[] Find $j$, $\mathcal{X}_j^{l}$, and $\mathcal{X}_j^{r}$ that minimize
                                \[
                                \mathcal{L}\left(
                                (y_i)_{i \in \mathcal{I}_{jl}},\delta_{\mathcal{I}_{jl}}
                                \right)
                                +
                                \mathcal{L}\left(
                                (y_i)_{i \in \mathcal{I}_{jr}},\delta_{\mathcal{I}_{jr}}
                                \right)
                                \]
                                \item[] such that $\left| \mathcal{I}_{jl}  \right| \geq n$, $\left| \mathcal{I}_{jr}  \right| \geq n$, $\mathcal{X}_j^l \cup \mathcal{X}_j^r = \mathcal{X}_j \cup \{\textrm{nan}\}$
                        \end{compactitem}
                        \item[] \textbf{Grow}
                                \begin{align*}
                                        h_l(x) & = \mathcal{T}\left(\left(y_i,x_i\right)_{\mathcal{I}_{jl}},d+1 \right) \\
                                        h_r(x) & = \mathcal{T}\left(\left(y_i,x_i\right)_{\mathcal{I}_{jr}},d+1 \right)
                                \end{align*}
                        \item[] \textbf{Output}
                                \begin{align*}
                                h(x) = \begin{cases}
                                h_l(x), & x_{\cdot j} \in \mathcal{X}_{j}^{l} \\
                                h_r(x), & x_{\cdot j} \in \mathcal{X}_{j}^{r}
                                \end{cases}
                                \end{align*}
                        \end{compactitem}
                        \end{compactitem}
                \end{flushleft}                
        \end{algorithm}
\newpage
        \begin{algorithm}[H]  
                \caption{Majority Rule training algorithm}
                \label{alg:majority}
                \begin{flushleft}
                        \textbf{Let}
                        \begin{compactitem}
                                \item $(y_i,x_i)_{i\in \mathcal{I}}$, where $y_i \in \mathcal{Y}$, $x_{ij} \in \mathcal{X}_j$, $j = 1,\ldots,p$, be the training data
                                \item $\mathcal{L}((y_i)_{i \in \mathcal{I}},\delta)$ be the loss given parameter $\delta$
                                \item $\delta_{\mathcal{I}}$ be the minimizing parameter of $\mathcal{L}((y_i)_{i \in \mathcal{I}},\delta)$
                                \item $\mathcal{I}_{jl} = \{i \in \mathcal{I}: x_{ij} \in \mathcal{X}_j^{l}\}$, $\mathcal{I}_{jr} = \{i \in \mathcal{I}: x_{ij} \in \mathcal{X}_j^{r}\}$, and $\mathcal{I}_{jm} = \{i: x_{ij} \textrm{ missing}\}$
                                \item $d_{\textrm{max}}$ be the maximum depth
                                \item $n$ be the minimum number of samples per node
                        \end{compactitem}
                        \textbf{Define training function} $\mathcal{T}\left((y_i,x_i)_{i \in \mathcal{I}},d\right)$ $\rightarrow$ $h(x)$:
                        \begin{compactitem}
                                \item[] \textbf{If} $d=d_{\textrm{max}}$ or $|\mathcal{I}| = n$:
                        \begin{compactitem}
                                \item[] \textbf{Output}
                                \begin{align*}
                                h(x) = \delta_\mathcal{I}
                                \end{align*}
                        \end{compactitem}
                        \item[] \textbf{Else}:
                        \begin{compactitem}
                                \item[] \textbf{Fit}
                        \begin{compactitem}
                                \item[] Find $j$, $\mathcal{X}_j^{l}$, and $\mathcal{X}_j^{r}$ that minimize
                                        \[
                                        \mathcal{L}\left(
                                        (y_i)_{i \in \mathcal{I}_{jl}},\delta_{\widetilde{\mathcal{I}}_l}
                                        \right)
                                        +
                                        \mathcal{L}\left(
                                        (y_i)_{i \in \mathcal{I}_{jr}},\delta_{\widetilde{\mathcal{I}}_r}
                                        \right)
                                        \]
                                \item[] where
                                \[
                                        \widetilde{\mathcal{I}}_l = \begin{cases}
                                        \mathcal{I}_{jl}, & \left| \mathcal{I}_{jl}  \right| \leq \left| \mathcal{I}_{jr}  \right| \\
                                                                        \mathcal{I}_{jl} \cup \mathcal{I}_{jm}, & \left| \mathcal{I}_{jl}  \right| > \left| \mathcal{I}_{jr}  \right|
                                        \end{cases}
                                \]
                                and vice versa for $\widetilde{\mathcal{I}}_r$, such that $\left| \widetilde{\mathcal{I}}_l  \right| \geq n$, $\left| \widetilde{\mathcal{I}}_r  \right| \geq n$, $\mathcal{X}_j^l \cup \mathcal{X}_j^r = \mathcal{X}_j$
                        \end{compactitem}
                        \item[] \textbf{Grow}
                                \begin{align*}
                                        h_l(x) & = \mathcal{T}\left(\left(y_i,x_i\right)_{\widetilde{\mathcal{I}}_l},d+1 \right) \\
                                        h_r(x) & = \mathcal{T}\left(\left(y_i,x_i\right)_{\widetilde{\mathcal{I}}_r},d+1 \right)
                                \end{align*}
                        \item[] \textbf{Output}
                                \begin{align*}
                                \hspace{-1.5cm}
                                h(x) = \begin{cases}
                                h_l(x), & x_{\cdot j} \in \mathcal{X}_{j}^{l} \textrm{ or } x_{\cdot j} \textrm{ missing and } \left| \mathcal{I}_{jl} \right| \geq \left| \mathcal{I}_{jr} \right| \\
                                h_r(x), & x_{\cdot j} \in \mathcal{X}_{j}^{r} \textrm{ or } x_{\cdot j} \textrm{ missing and } \left| \mathcal{I}_{jl} \right| < \left| \mathcal{I}_{jr} \right|
                                \end{cases}
                                \end{align*}
                        \end{compactitem}
                        \end{compactitem}
                \end{flushleft}                
        \end{algorithm}
        
        \begin{algorithm}[H] 
                \caption{Fractional Case training algorithm}
                \label{alg:fc}
                \begin{flushleft}
                        \textbf{Let}
                        \begin{compactitem}
                        \item $(y_i,x_i,w_i)_{i\in \mathcal{I}}$, where $y_i \in \mathcal{Y}$, $x_{ij} \in \mathcal{X}_j$, $j = 1,\ldots,p$, $w_i\in[0,1]$ be the training data
                        \item $\mathcal{L}((y_i,w_i)_{i \in \mathcal{I}},\delta)$ be the loss given parameter $\delta$
                        \item $\delta_{\mathcal{I}}$ be the minimizing parameter of $\mathcal{L}((y_i)_{i \in \mathcal{I}},\delta)$
                        \item $\mathcal{I}_{jm} = \{i \in \mathcal{I}: x_{ij} \textrm{ missing}\}$, $\mathcal{I}_{jl} = \{i \in \mathcal{I}: x_{ij} \in \mathcal{X}_j^{l}\} \cup \mathcal{I}_{jm}$, $\mathcal{I}_{jr} = \{i: x_{ij} \in \mathcal{X}_j^{r}\} \cup \mathcal{I}_{jm}$
                        \item $d_{\textrm{max}}$ be the maximum depth
                        \item $n$ be the minimum total sample weight per node
                        \end{compactitem}
                        \textbf{Define training function} $\mathcal{T}\left((y_i,x_i)_{i \in \mathcal{I}},d\right)$ $\rightarrow$ $h(x)$:
                        \begin{compactitem}
                        \item[] \textbf{If} $d=d_{\textrm{max}}$ or $|\mathcal{I}| \leq n$:
                        \begin{compactitem}
                                \item[] \textbf{Output}
                                \begin{align*}
                                h(x) = \delta_\mathcal{I}
                                \end{align*}
                        \end{compactitem}
                        \item[] \textbf{Else}:
                        \begin{compactitem}
                                \item[] \textbf{Fit}
                        \begin{compactitem}
                        \item[] Find $j$, $\mathcal{X}_j^{l}$, and $\mathcal{X}_j^{r}$ that minimize
                                \[
                                \mathcal{L}\left(
                                (y_i,w_i^l)_{i \in \mathcal{I}_{jl}},\delta_{\mathcal{I}_{jl}}
                                \right)
                                +
                                \mathcal{L}\left(
                                (y_i,w_i^r)_{i \in \mathcal{I}_{jr}},\delta_{\mathcal{I}_{jr}}
                                \right)
                                \]
                                \item[] where
                        \[
                                w_i^{l}  = \begin{cases}
                                w_i, & x_{ij} \in \mathcal{X}_j^l \\
                                w_i\frac{\left|
                                \mathcal{I}_{jl}\setminus\mathcal{I}_{jm}
                                \right|
                                }{
                        \left|
                                \mathcal{I}\setminus\mathcal{I}_{jm}
                                \right|
                                }, & x_{ij} \in \{\texttt{nan}\}
                                \end{cases}
                        \]
                        such that $\sum\limits_{i \in \mathcal{I}_{jl}} w_i^l \geq n$, and vice versa for $r$, and $\mathcal{X}_j^l \cup \mathcal{X}_j^r = \mathcal{X}_j$
                        \end{compactitem}
                        \item[] \textbf{Grow}
                                \begin{align*}
                                        h_l(x) & = \mathcal{T}\left(\left(y_i,x_i,w_i^l\right)_{\mathcal{I}_{jl}},d+1 \right) \\
                                        h_r(x) & = \mathcal{T}\left(\left(y_i,x_i,w_i^r\right)_{\mathcal{I}_{jr}},d+1 \right)
                                \end{align*}
                        \item[] \textbf{Output}
                                \begin{align*}
                                h(x) = \begin{cases}
                                h_l(x), & x_{\cdot j} \in \mathcal{X}_{j}^{l} \\
                                h_r(x), & x_{\cdot j} \in \mathcal{X}_{j}^{r} \\
                                w_l h_l(x) + w_r h_r(x), & x_{\cdot j} \textrm{ missing}
                                \end{cases}
                                \end{align*}
                        \end{compactitem}
                        \end{compactitem}
                        \end{flushleft}                
        \end{algorithm}
        
\end{multicols}
\end{document}